\def\year{2020}\relax
\documentclass[letterpaper]{article} % DO NOT CHANGE THIS
\usepackage{aaai20}  % DO NOT CHANGE THIS
\usepackage{times}  % DO NOT CHANGE THIS
\usepackage{helvet} % DO NOT CHANGE THIS
\usepackage{courier}  % DO NOT CHANGE THIS
\usepackage[hyphens]{url}  % DO NOT CHANGE THIS
\usepackage{graphicx} % DO NOT CHANGE THIS
\usepackage{graphicx}  % DO NOT CHANGE THIS
\title{Automatic Building and Labeling of HD Maps with Deep Learning}

\author{Mahdi Elhousni, Yecheng Lyu, Ziming Zhang, Xinming Huang
\\
Worcester Polytechnic Institute
\\
\{melhousni, ylyu, zzhang15, xhuang\}@wpi.edu}

\begin{document}
%
%\frenchspacing
\maketitle

\begin{abstract}
In a world where autonomous driving cars are becoming increasingly more common, creating an adequate infrastructure for this new technology is essential. This includes building and labeling high-definition (HD) maps accurately and efficiently. Today, the process of creating HD maps requires a lot of human input, which takes time and is prone to errors. In this paper, we propose a novel method capable of generating labelled HD maps from raw sensor data. We implemented and tested our methods on several urban scenarios using data collected from our test vehicle. The results show that the proposed deep learning based method can produce highly accurate HD maps. This approach speeds up the process of building and labeling HD maps, which can make meaningful contribution to the deployment of autonomous vehicles.
\end{abstract}

\section{I - Introduction}
During the last couple of decades, autonomous driving has become an important research topic in the scientific community. This stems from the fact that scientists, governments and people in general are starting to realise the huge positive impacts that autonomous driving could have on our daily lives. According to a report by The Department of Transportation of the USA, self-driving cars could reduce traffic fatalities by up to 94\% by eliminating the accidents that are due to human errors.\par
The race toward fully autonomous driving cars, or level 5 autonomy as categorized by SAE International, has given rise to multiple new fields and was the catalyst to launch or greatly improve several new disciplines. This manifested itself in the field of deep learning, which has made it possible today to achieve a respectable level of autonomy when driving on roads that fall into the classic scenario box. Lanes and roads (or driveable regions) detection can be achieved with neural networks trained to excel in task related to pixel-wise segmentation of images captured by cameras \cite{long2015fully}, making the car aware of where it is safe to drive. Other types of networks are trained to detect obstacles and classify them into several independent classes \cite{redmon2017yolo9000}, sometimes with the help other sensors such as radars to mitigate the accuracy issue when it comes to depth and 2D images. This helps the car drive safely by avoiding obstacles on the road and obeying traffic rules represented by traffic signs or traffic lights. \par

This deep learning approach has multiple advantages that revolve mostly around the real-time aspect and the use of low-cost sensors. However, methods based solely on deep learning and cameras (while being very performant in highway scenarios for example) are bound to fail when deployed in urban scenarios because cameras have major weaknesses, mostly related to brightness issues. These weakness could be covered by fusing the camera data with other more powerful and accurate sensors, such as LiDARs. This introduces the most accurate method to date to navigate and drive autonomously in urban areas, which is based on HD maps. \par
HD maps are a combination of 3D point clouds and relevant semantic information. 3D point clouds can be used for localization by matching the incoming scans of the LiDAR when driving, with the pre-build and stored 3D point clouds. However, in order for these point clouds to be used for autonomous navigation, additional data has to be stacked on top of it. Information such as the position of the lanes, roads or traffic signs has to be labelled in order for the car to navigate safely while obeying basic traffic rules. The usual solution used today to incorporate this information on the maps is to manually label them and store them. An obvious shortcomings of this method is the lack of accuracy. Since point clouds do not have RGB values which makes lanes and lane markings invisible, and the use of the intensity field of the pointcloud can lead to very noisy data. Also labeling point clouds is known to be very tedious and can slow down the HD maps building process significantly. \par
Alternatively we can use the detection results generated by neural networks to build and label HD maps offline before deploying them on cars. Therefore, we propose in this paper to use deep learning to automate the labeling process of HD maps, by combining them with other methods to improve the accuracy and robustness.  Our contributions can be summarized as a collection of algorithms and pipelines aiming to automatically build and label HD maps for urban autonomous driving. \par
This paper is organized as follows : Section II introduces some of the related work in building and labeling HD maps. Section III presents our pipeline for building HD maps and both of the pipelines for road and lanes labelling; in Section IV we present the results that we obtained in different real world scenarios. Finally, Section V concludes this work.

\section{II - Related Work}
Used by many major companies in the autonomous driving business and research community such as Waymo or Autoware \cite{kato2015open}, navigation using HD maps has proven to be the most robust method up to date. HD maps are maps based on laser data collected using a LiDAR sensor. The most recent HD maps are built using a 360 degrees  rotating LiDAR sensor, where after each full rotation, a scan is packaged and sent to the computer. Multiple scans are then accumulated in order to generate a point cloud where the 3D geometry of the surrounding environment is represented. Accumulating these scans can be done using multiple approaches : The Iterative Closest Point (ICP) algorithm \cite{zhang1994iterative} is a popular algorithm designed to minimize the difference between two point clouds and thus finding the right transformation between them. ICP was the leading and most robust algorithm to align two consecutive point clouds until it was surpassed by the 3D Normal Distribution Transform (3D NDT) algorithm \cite{magnusson2007scan}. \par
The 3D NDT algorithm was proposed by M. Magnusson et al. in order to help the deployment of autonomous mining vehicles. It builds on the 2D method of the same name \cite{biber2003normal} by transforming the point cloud into a probability density function which can be used with Newton's algorithm to match another point cloud. In his paper, M. Magnusson shows that the 3D NDT algorithm outperforms the ICP algorithm, which was the standard 3D matching algorithm at that time. In order to improve both the speed and accuracy of the 3D NDT algorithm, an EKF is used to fuse both IMU and GPS data to generate an initial guess of the transformation that is passed to the 3D NDT algorithm as a starting point for the minimization procedure. \par
After constructing the point cloud, it is necessary to add semantic information to it such as lanes and driveables regions in order for the car to be able to autonomously navigate itself. Traditionally, this information is labelled on the point cloud using a manual method. As an example, Autoware gives access to its users to an online tool named Vector Mapper, where they can load up their pre-constructed point clouds and label them manually. The labels are then exported and available for download as CSV files, called Vector Maps. However, manual labeling of point clouds tends to be extremely tedious, time-consuming and not as accurate as we would like it to be. Also what could be an one-step process is now divided into two independent ones where the point cloud is built as first step then manually labeled as a second step. Since most of the information that is being labelled can be generated from deep learning neural networks, we propose in this paper to bypass the manual labeling process and use the predictions generated by the networks to label the point cloud automatically. We also propose other post-predictions methods to mitigate the false positives and uncertainty that tend to occur sometimes when using deep learning. \par
The deep learning networks that we deploy in this process are monocular camera based and not LiDAR based. This choice stems from the fact that the level of accuracy reached by camera based networks has yet to be matched by the LiDAR based ones and also by the fact that deploying multiple LiDAR networks at the same time will require a lot more processing power. The LiDAR might outperform the camera when it comes to range, but our algorithm here relays on accumulating information from successive samples which helps covering a larger amount of space. In order to fuse both the camera predictions and the LiDAR point clouds, a transformation consisting of a translation and rotation between the camera and LiDAR is needed, and can be found using a lidar-camera calibration method. One of the proposed methods in the literature, which we used in this paper, is described in \cite{lyu2019interactive}.

\section{III - Methods}

\subsection{Mapping pipeline } The mapping pipeline presented in this paper concentrates on constructing the 3D geometry of the surrounding environment while labeling the road information on the HD map. Here, the road information that we label are the driveable regions and the lanes. Fig. 1 shows an overview of the full pipeline. For clarity purposes, we define three frames : The map frame $F_m$, whose origin is the center of the first scan at the start of the map, the car frame $F_c$, whose origin is the center of gravity of the autonomous car and the LiDAR frame $F_l$ whose origin is the center of the LiDAR sensor.

\begin{figure}[h]
\centering
\includegraphics[width=0.86\columnwidth]{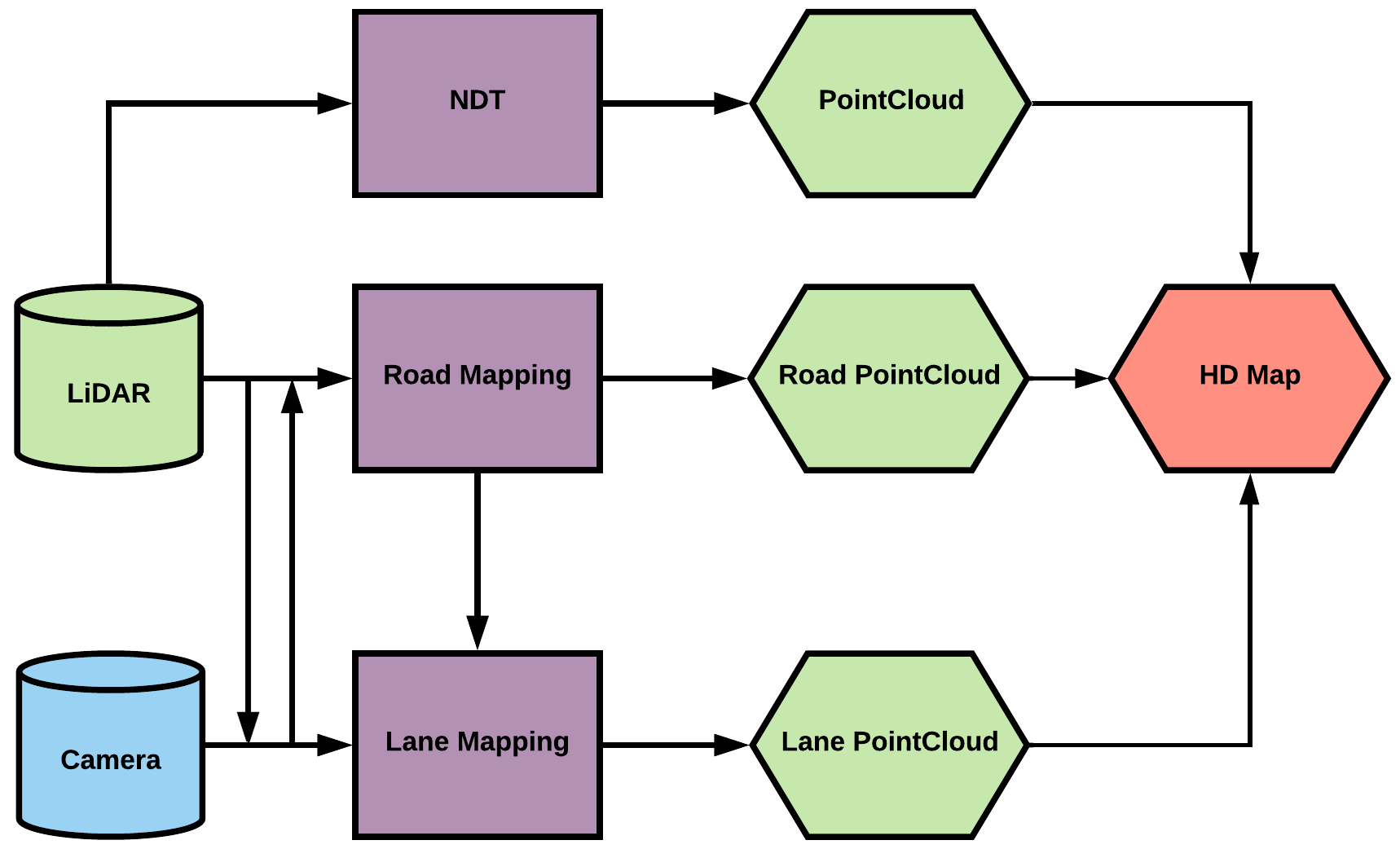}
\caption{Mapping pipeline combines 3D-NDT with pre-trained DNN to generate labeled HD maps.}
\end{figure}

\subsection{Road mapping } We define a road $R$ as polygon in the $F_m$ frame limiting the areas where it is possible to drive, but not necessarily legal to do so. Road mapping is performed for two main reasons :
\begin{itemize}
    \item It can guide the autonomous car when no lanes are present on the road, as it is sometimes the case for one way streets.
    \item It can assist the lane mapping that we will describe later.
\end{itemize}

The road is detected using the camera data, projected on the LiDAR data, refined to remove outliers and then accumulated with the previous scans using the output of the 3D NDT algorithm. We then compute the region occupied by the road and extract the road limits. We will explain each of these steps in the following. Fig. 2 shows an overview of the road mapping pipeline. \newline

\begin{figure}[h]
\centerline{\includegraphics[width=0.9\columnwidth]{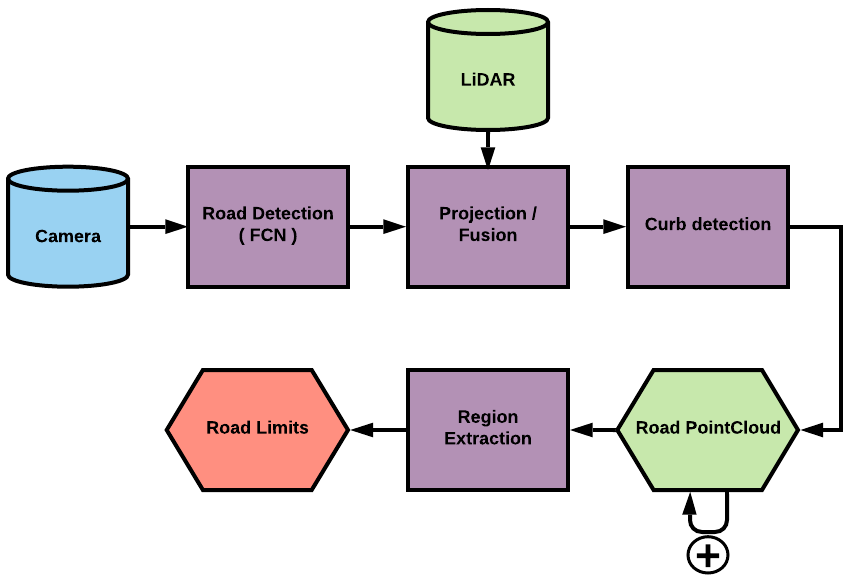}}
\caption{Road mapping pipeline. The results from the camera FCN are pruned to remove outliers.}
\end{figure}

\noindent
\textbf{Detection} : For detecting the road we use a Fully Convolutional Network (FCN) \cite{long2015fully}. We apply the network to the front camera data, in order to segment the image into 2 regions : Road and Not Road. This results into a binary image that we will use in combination with the Lidar-Camera calibration to segment the point cloud of the road. \newline
\textbf{Projection} : We start by using the camera parameters to crop the 360 point cloud, so that we are only operating on the points that fall within the field of view of the camera. Then, using the Lidar-Camera calibration, we project the binary image on top of the point cloud while making sure that the color information from the image is preserved and transferred to the point cloud. This results into a binary point cloud where the road points are colorized differently than other points on the point cloud. \newline
\textbf{Curb detection} : In some cases the road detected by the FCN tends to bleed on the edges of the curb, especially when the curb is hard to distinguish in the image frames because of shadows, brightness changes or the curb being too small. Therefore, we need to refine the results from the FCN by removing any curb portions that were included in the predicted road.

        \begin{figure}
        \centering
        \includegraphics[width=\columnwidth]{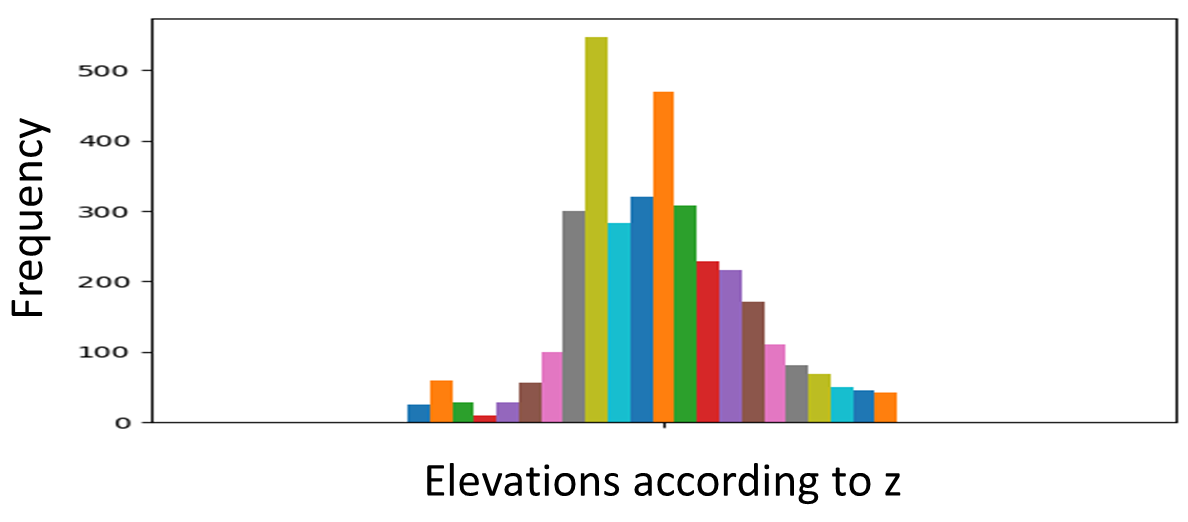}
        \caption{Histogram of the elevation z of the road point cloud.}
        \label{fig:binomial fig}
        \end{figure}

In order to do so, we use the colorized point cloud and the elevation according to the z axis. We start by extracting the road point cloud from the colorized point cloud that we obtained previously using a color segmentation based method, and then, as it is shown in Fig. 3, we display the elevation of the points in the road point cloud as a histogram. It shows us that the points in the extracted road point cloud follow a bimodal distribution, meaning a distribution that contains two peaks representing the two normal distributions with means $\mu_1$ and $\mu_2$ respectively and standard deviations $\sigma_1$ and $\sigma_2$ respectively. This makes sense because the first normal distribution represents the points on the curb, and the second one represents the points on the road.
In this case, detecting the curb consists of separating the bimodal distribution into two normal distributions and excluding the distribution that contains the curb points. To achieve that, we use a method commonly employed in computer vision for segmentation and clustering purposes called the Otsu method \cite{otsu1979threshold}. \par The Otsu method, applied to a bimodal distribution, calculates the optimum threshold separating the two classes (in our case “road” and “curb”). This makes it possible to exclude most of the points that lay on the curb as it shown on Fig 4 and leaves us with a portion of points, which the elevation of, follows a normal distribution with a mean $\mu_1$ and standard deviation $\sigma_1$. As a final check, and to remove the rest of the outliers, we apply the 68--95--99.7 rule to the resulting distribution and exclude all the points which the elevation lay outside of $\mu_1 - \sigma_1$ and $\mu_1 + \sigma_1$.

    \begin{figure}[h]
        \centering
        {\includegraphics[width=\columnwidth]{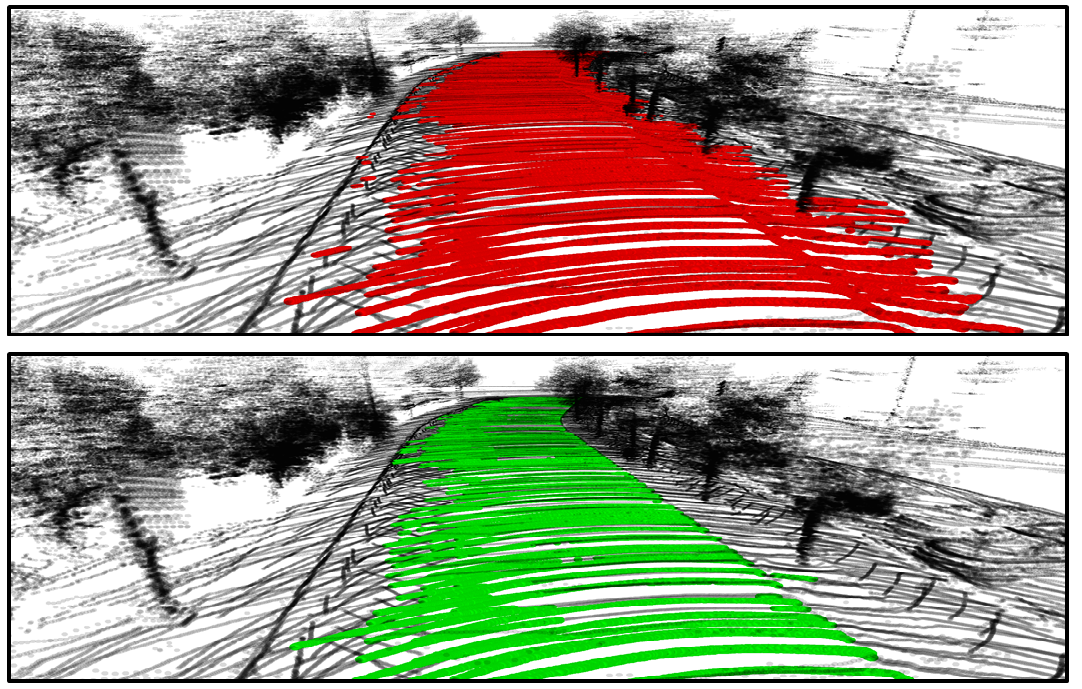}}
        \caption{Road mapping before (Red) and after (Green) the curb detection.}
    \end{figure}

\noindent
\textbf{Region Extraction} : In order to extract the limits of the driveable area, we need to compute the contour of the road point cloud projected onto the $(x,y)$ plane. This is achievable by using a Concave Hull (CH) \cite{moreira2007concave} which is an algorithm based on the k-nearest  neighbours approach and designed to generate a envelope describing the area occupied by a set of points in a plane. The envelope generated by the CH is used to contruct the polygon describing the driveable area.

\subsection{Lane mapping} We define a lane $L$ as a set of points $L = \{p_1, p_2, ..., p_n\}$ where $p_i = \{x_i, y_i, z_i\}$ are the coordinates of the $i'th$ point in the $F_m$ frame. Lane mapping is performed in order to help the autonomous car navigation process on the road by keeping it centered. The lanes are detected using the camera data, projected on the LiDAR data, clustered and smoothed to generate meaningful waypoints and then accumulated with the previous scans using the output of the 3D NDT algorithm. We also generate the missing lanes that were not detected in the camera data. We will explain each of these steps in the following. Fig. 5 shows an overview of the lane mapping pipeline.  \newline

\begin{figure}[hb]
\centerline{\includegraphics[width=0.9\columnwidth]{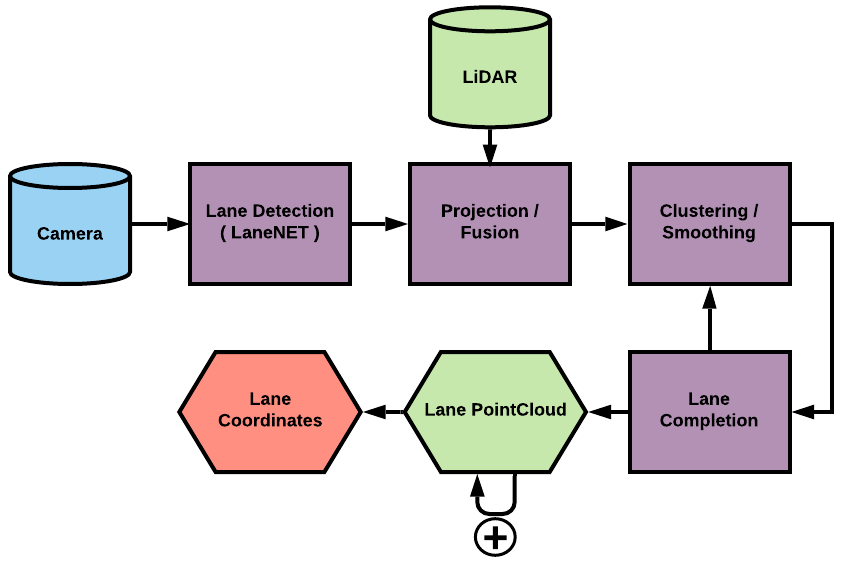}}
\caption{Lane mapping pipeline. The results from the camera LaneNET are smoothed and clustered to generate lane waypoints.}
\end{figure}

\noindent
\textbf{Detection} : For lane detection, we use LaneNET \cite{wang2018lanenet}. This network was selected because it is able to detect all the lanes that are visible from the front view camera and not only the ego lanes. The network outputs a mask image of the same size as the input image, where the pixels belonging to the lanes are labeled and color coded. Similarly to what we did for the road mapping, the mask image will be combined with the lidar-camera calibration to generate a lanes point cloud. \newline
\textbf{Projection} : Since the lidar-camera calibration will become less accurate the further we are from car, we start by cropping the ``camera field of view point cloud" to a certain distance L from the origin of the $F_l$ frame before projecting the lanes mask on it. This helps preserving the shape of the lanes as we will be accumulating the projections and point clouds while advancing. Finally, using a color segmentation method, we extract the points belonging to the lanes to form a lanes point cloud. \newline
\textbf{Clustering \& smoothing} : The generated lanes pointcloud is noisy and does not always follow a coherent geometry. Thus we set up a series of clustering and smoothing steps that will be applied to the lanes point cloud in order to generate a series of waypoints that can be used by the autonomous car to know the positions of the lanes in the space. The smoothing and clustering is applied on two different levels : first in the $F_l$ frame when dealing with a single scan, then secondly in the $F_m$ frame when the current scan has been accumulated with the previous ones using the output of the 3D NDT algorithm. \par We first start, in the $F_l$ frame, by dividing the resulting lanes point cloud into equally separated regions of interests so we can operate on the different lanes independently. These lanes are then clustered following the method proposed in \cite{rusu2010semantic} which is mainly based on the Euclidean distance between neighboring points, and which generate points that describe the geometry and curvature of the lane. These points are then smoothed by being fitted to quadratic function before being transformed into the $F_m$ frame and accumulated with the previous scans using the result of the 3D NDT algorithm. When in the $F_m$ frame, the clustering follows the same method used in the first step, but with a higher tolerance and the smoothing is done by fitting the points to a logarithmic curve instead of a quadratic one. \par
We define the smoothing index $N$ as the index in the lane $L$ from where the smoothing process starts and the “look-back” parameter $l$ that defines how much of the full lane do we want to include in the smoothing process. The smoothing and clustering of the lane points will be applied if certain criterias are met :

     \begin{itemize}
     \item If the accumulated distance of the lane portion is larger than a threshold $L1$, then the lane portion is smoothed and the smoothing index is moved to the $N’th$ position, where $N = n - m - l$, $n$ being the size of the full lane point cloud $L$ and $m$ the size of the last lane portion $L' = \{p_{n+m}, p_{n-m+1}, ..., p_n\}$ that was added during the previous scan.
     \item If we receive $S$ scans without lane points, signaling the presence of an intersection for example, that lane portion is smoothed and the smoothing index is moved to the $n'th$ position, where $n$ is the size of the full lane point cloud.
     \item  If the accumulated distance of the lane portion is larger than a threshold $L2$, that lane portion is clustered in order to produce reference points.
     \end{itemize}

Setting up the thresholds $L1$, $L2$ and $l$ is critical in order to obtain lane points that are not too curved/straight or too dense/light.

\noindent
\textbf{Lane completion} : LaneNET sometimes fails to detect a lane either because of a sudden change in brightness/contrast, the lane not being in the field of view of the camera or the lane simply not being drawn on the road. We deal with this issue by combining the successful lanes that were detected, our curb detection algorithm and the fact that the lanes on the road are parallel. \newline
We first start by using the curb detection results to check if all the lanes were detected : based on the position of the curbs and the lanes width (which is derived from the successful detections), we can conclude if the right number of lanes was detected. If a certain lane is missing, we use the closest left or right lane to it to generate it by fitting the lane points obtained from the last scan to a quadratic curve and then combining the normals to the curve at each of those points with the lanes width to generate a new lane.

\section{IV - Results }

\subsection{Experimental setup :}
Our experimental setup consists of a Lincoln MKZ, equipped with one Velodyne Lidar VLP-16 and one FLIR PointGrey RGB camera recording at 10 Hz and 30 Hz respectively. The car is also equipped with an IMU/GPS to assist the 3D NDT algorithm. The data is synchronized and recorded throught the Robot Operating System (ROS) as ROS bags and then processed offline.
We recorded and built multiple HD maps using the pipeline, from which we selected 5 scenarios.
Each of these scenarios was selected for a specific reason :

\begin{itemize}
    \item \textbf{$straightRoad$} was selected to demonstrate the effectiveness of the curb detection step in the road mapping pipeline.
    \item \textbf{$curvedRoad$} was selected to demonstrate the accuracy of the Smoothing and Clustering step in the lane mapping pipeline.
    \item \textbf{$mergeLane$} was selected to demonstrate that the lane mapping pipeline is able to handle scenarios where new lanes appear and that we are not restricted to the lanes that we started with.
    \item \textbf{$intersection$} was selected to demonstrate the results of merging multiple maps.
    \item \textbf{$highway$} was selected to demonstrate that the mapping pipeline is still effective in significantly larger areas.
\end{itemize}

The ground truth was manually labeled using the intensity of the LiDAR point cloud and satellite imagery as reference. When building and labeling our HD maps using our automatic pipeline, we set our thresholds such as $L1 = L2 = 31$ meters, $l = 5$ and $S = 5$.

\begin{figure*}[t]
\centering
\includegraphics[width=\linewidth]{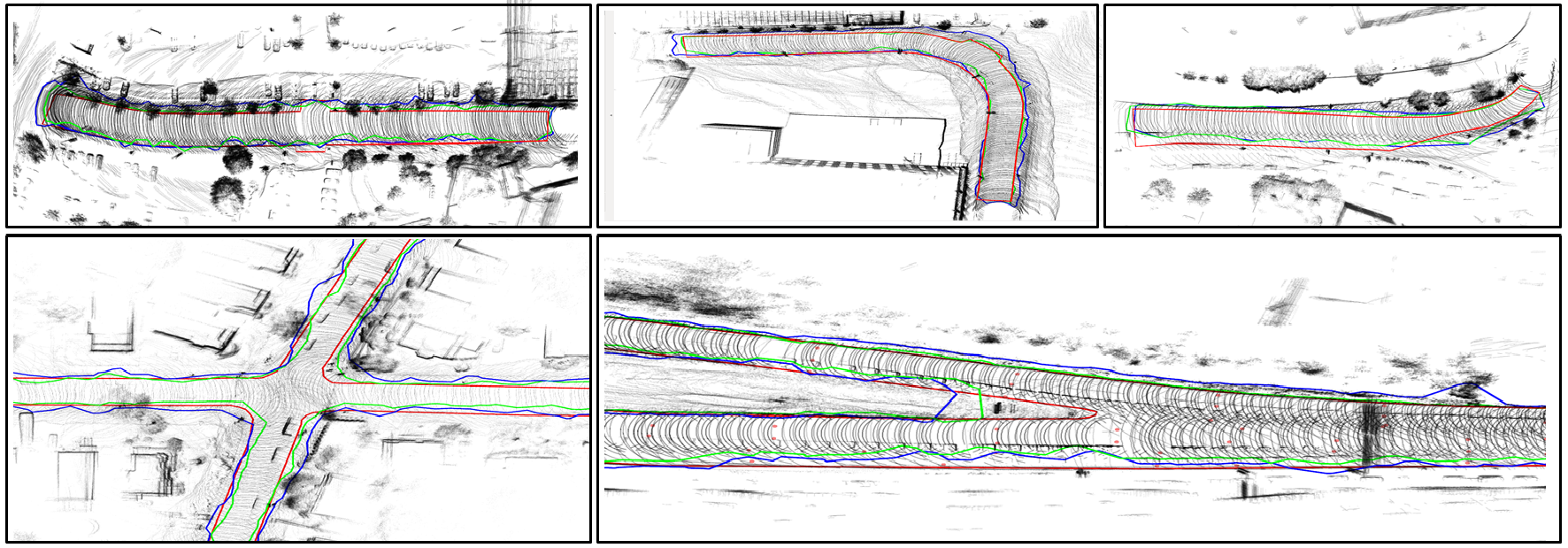}
    \caption{Qualitative comparison : Red is the ground truth, Blue is before the curb detection, and Green is after.}

\centering
\includegraphics[width=\linewidth]{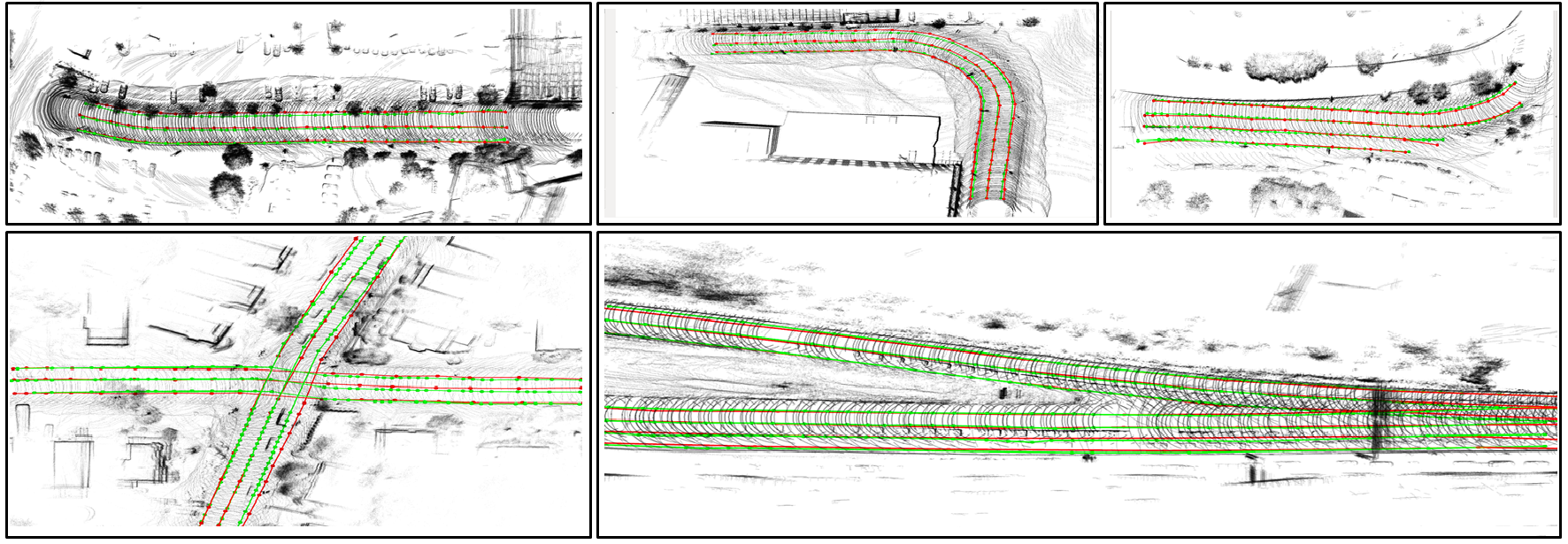}
    \caption{Qualitative comparison : Red lanes are the ground truth and Green ones are automatically generated.}
\end{figure*}

\subsection{Road Mapping :}

We present the results of our Road Mapping pipeline : Fig 6 shows a qualitative comparison between the manually labeled ground truth and automatically labeled roads, with and without the curb detection algorithm. Table 1 lists the errors in the areas occupied by the road polygon before and after the curb detection algorithm. The results show that the curb detection algorithm is very effective in excluding the points that do not belong to the road. However, traffic on the road can sometimes lead to results where valid road points are excluded, as it is the case during the $mergeLane$ recording. A potential solution would be to detect the objects on the road and exclude the points that belong to those object before applying the curb detection algorithm.

\begin{table}[h]
	\centering
    \caption{Errors ($m^2$) in the areas occupied by the mapped road. $\delta$ represents the percentage of points that were exculed.}
    \begin{tabular}{|c|c|c|c|}
      \hline
        & {Before CD} & {After CD} & {$\delta$}\\
      \hline
      $straighRoad$ & 0.232 & 0.008 & 0.16 \\
      \hline
      $curvedRoad$ & 0.261 & 0.034 & 0.22 \\
      \hline
      $mergeLane$ & 0.252 & 0.127 & 0.05 \\
      \hline
      $intersection$ & 0.357 & 0.017 & 0.18\\
      \hline
      $highway$ & 0.007 & 0.160 & 0.32\\
      \hline
    \end{tabular}
\end{table}

\subsection{Lane Mapping :}

We also present the results of our Lane Mapping pipeline : Fig 7 shows a qualitative comparison between the manually labeled ground truth and automatically labeled lanes. Table 2 lists the translation error mean values, while Table 3 lists the standard deviations in $x$ and $y$ of the automatically labelled lanes from the ground truth. When being evaluated, both the maps generated from the $intersections$ and $highway$ were split into two maps representing each of the roads present. These results show that the Lane Mapping pipeline is capable of accurately labeling the lanes and generating the missing ones. However, this latter part is highly dependent on the success of the road mapping pipeline in finding the furthest curb.

\begin{table}[h]
	\centering
    \caption{Mean values of the standards deviations of the automatically labeled lanes from the ground truth according to $x$ and $y$.}
    \begin{tabular}{|c|c|c|}
      \hline
        & \textbf{$\sigma_x$} & \textbf{$\sigma_y$}\\
      \hline
      $straighRoad$ & 0.522 & 0.206\\
      \hline
      $curvedRoad$ & 0.324 & 0.508\\
      \hline
      $mergeLane$ & 0.481 & 0.444\\
      \hline
      $intersection_1$ & 0.384 & 0.303\\
      \hline
      $intersection_2$ & 0.634 & 1.101\\
      \hline
      $highway_1$ & 0.613 & 0.619\\
      \hline
      $highway_2$ & 0.626 & 0.340\\
      \hline
    \end{tabular}
\end{table}

\begin{table}[h]
  \begin{center}
    \caption{Translation error (m) between the automatically labeled lanes and the ground truth.}
    \begin{tabular}{|c|c|c|c|c|}
      \hline
        & \textbf{$Lane_1$} & \textbf{$Lane_2$} & {$Lane_3$} & {$Lane_4$}\\
      \hline
      $straighRoad$ & 0.918 & 0.605 & 0.300 & N/A\\
      \hline
      $curvedRoad$ & 0.746 & 0.586 & 1.311 & N/A\\
      \hline
      $mergeLane$ & 0.764 & 0.779 & 0.590 & 0.877\\
      \hline
      $intersection_1$ & 1.238 & 0.409 & 0.583 & N/A\\
      \hline
      $intersection_2$ & 0.635 & 0.951 & 1.054 & N/A\\
      \hline
      $highway_1$ & 0.507 & 0.625 & 0.764 & N/A\\
      \hline
      $highway_2$ & 0.585 & 0.583 & 0.618 & 0.565\\
      \hline
    \end{tabular}

  \end{center}
\end{table}

\section{V - Conclusion}
In this paper, we proposed a pipeline designed to automatically build and label HD maps for autonomous driving cars. The pipeline relies on the results of deep learning networks trained to detect the driveable areas and the lanes. These results are then automatically post-processed in order to be corrected, improved or completed. A comparison between the results of our pipeline and a manually labelled ground truth proved the accuracy and effectiveness of the methods employed in this work. Future work includes automatic labeling of more details on the HD maps such as traffic sign and traffic lights.

\bibliographystyle{aaai}
\bibliography{bibfile1}

\end{document}